**Phenology-based learning framework for yield estimation and harvest forecasting of raspberry fruits**

Parham Jafary[1], Lesley G. Campbell[2], Anna Bazangeya[2], Michelle Pham[2], Sajad Saeedi[3], Kourosh Zareinia[1], Habiba Bougherara[1*]

[1] Dept. Of Mechanical, Industrial and Mechatronics Eng., Toronto Metropolitan University, Toronto, Ontario, Canada

[2] Dept. Of Chemistry and Biology, Toronto Metropolitan University, Toronto, Ontario, Canada

[3] Dept. Of Computer Science, University College London, United Kingdom

[*] Corresponding Author: habiba.bougherara@torontomu.ca

**Abstract:**

The future of agriculture is intertwined with automation. Accurate fruit detection, yield estimation, and harvest time prediction are crucial for efficient supply chain management by optimizing resources and logistic utilization. Computer vision can automate these tasks to reduce labour costs and improve efficiency by training deep learning models on appropriate data to perform knowledge-based tasks. Although fruit detection has been the focus of literature, yield prediction and harvest time estimation remain practical challenges for farmers. This is particularly important for high-value, highly perishable crops, such as raspberries, where contractual obligations require precise harvest timing. This paper addresses this gap by providing a learning-based framework for raspberry maturity detection and harvest time estimation. For this end, a phenology study of raspberry plants from three different cultivars was carried out, and seven development stages were identified based on the BBCH-scale; for the first time, a phenology-based dataset of developing raspberry flowers and fruit was curated and made available publicly, which contains 1,853 high-resolution images and 6,907 manually labelled annotations. A comprehensive benchmark was developed from state-of-the-art object detection models, and finally, a tailored deep learning model capable of real-time inference was established that achieved 92.2% detection accuracy in the vertical farm field test. This paper provides a model that can estimate raspberry yield 28 to 33 days in advance of harvesting, a dataset that can be used by other researchers for object detection and segmentation model development, and a framework that can be extended for precise yield and harvest time estimation.

**Nomenclature**

| Abbreviation | Definition |
|---|---|
| AP | Average Precision |
| BB | Bounding Box |
| BBCH | Biologische Bundesanstalt, Bundessortenamt und CHemische Industrie |
| DL | Deep Learning |
| DNN | Deep Neural Networks |
| FPS | Frames Per Second |
| FN | False Negative |
| FP | False Positive |
| GPU | Graphical Processing Unit |
| IoU | Intersection over Union |
| mAP | Mean Average Precision |
| MP | Megapixel |
| NeRF | Neural Radiance Fields |
| P | Precision |
| R | Recall |
| RAM | Random Access Memory |
| RGB-D | Red, Green, Blue – Depth |
| RT-DETR | Real-Time Detection Transformer |
| SD | Standard Deviation |
| TP | True Positive |
| YOLOv* | You Only Look Once version * |

# 1 Introduction

Raspberries are a high-value berry crop with well-documented nutritional benefits [1–5], that make them appealing to both growers and consumers. They are grown in many countries, with summer-bearing varieties being common in colder regions (e.g., Canada, Russia, Poland) and ever-bearing varieties being common in milder climates (e.g., California, Mexico, Chile). However, their high perishability (2-5 days post-harvest) [6, 7] makes harvest timing and supply chain management particularly challenging for growers. Moreover, accurate in-season yield and harvest-time estimation are crucial for growers, as they risk financial penalties for under- and over-delivery of perishable goods due to contractual obligations with retailers and distributors. It also improves pricing stability, market position, and logistical planning [8]. To achieve accurate yield estimation, growers need to have a deep understanding of the speed of fruit or vegetable ripening and the developmental stages of the fruit, patterns often captured by phenology studies [9–11]. Phenological monitoring, such as recording the timing of budding, flowering, and fruiting, provides critical insights into crop development. This data enables precision agriculture practices, improving decisions on irrigation, nutrient application, labour and logistics planning.

The use of non-invasive technologies for early assessment of fruit ripeness metrics is increasingly regarded as a cornerstone for managing harvest logistics, ensuring fruit quality, and mitigating waste [12]. Current methods, in raspberry cultivation, rely on manual tracking of fruit maturity, a labour-intensive and error-prone process highly dependent on empirical knowledge. Thus, automating fruit monitoring, harvest time, and yield estimation would reduce labour costs and food waste, while increasing profitability [13]. Automation in agriculture is intertwined with the use of remote sensing techniques. While the use of spectroscopy is well-explored for agricultural practices [14–18], recent trends have shifted towards computer vision systems based on deep learning (DL) models [19]. For instance, Bazame et al. used the YOLOv4 model to detect and count coffee fruits from harvest videos and achieved a precision of 83.5% [20]. Sa et al. used Faster R-CNN to distinguish among fruits, achieving an F1-score of 0.83 across all classes [21]. Other researchers have employed DNNs to characterize fruit ripeness [22, 23]. Arad et al. developed an autonomous robot that used RGB-D images to harvest sweet peppers [24]. Fujinaga et al. addressed the need for multifunctional agricultural robots in small-scale farming by developing a method for detecting cutting points for strawberry harvesting and truss pruning [25]. Zhang et al. introduced YOLOMS, a multi-task learning model based on an improved YOLOv5s architecture, for mango recognition and stem picking point localization in natural environments, achieving a mean average precision (mAP) of 82.42% and a recall of 85.64% [26].

The main contributing factors for shifting from spectrometers to cameras are the lower cost of a camera, less complexity of the integrated system, and more interpretability of images than reflectance readings. However, the downside of DL models is that they require large datasets for the training, validation, and testing process. More importantly, to produce practical results, the data needs to be labelled, which adds to the complexity of the data preprocessing and makes the dataset development task a tedious and time-consuming process. These facts underscore the importance and value of publicly available datasets.

To help address this problem, general-purpose datasets such as COCO [27] and LVIS [28] were developed and are often used for pretraining models, but they lack relevance for agricultural contexts, offering neither appropriate environmental conditions nor crop-specific targets. As a result, models trained on these datasets require fine-tuning with task-specific data (a process known as transfer learning) to be applicable in agricultural practices such as fruit detection or harvest time estimation. Researchers have widely leveraged transfer learning and fine-tuned networks that were pretrained on a large dataset using a relatively small dataset [29–33]. Even when transfer learning is used, performance is limited by the availability and quality of agricultural datasets [34]. Fruit-specific datasets have been developed for apples, strawberries, tomatoes, grapes, and cherries [35–39], and are commonly used to train object detection models such as YOLO, RT-DETR, or Faster R-CNN. However, only a few are publicly available, they have only a few hundred images, or their images have low resolution, all of which prevent achieving robust DL models [40].

Deep learning-based fruit detection and classification have improved in recent studies, but most of these studies use simple labels such as “immature”, “ripe” or “damaged” in their datasets [23, 38, 41, 42]. These labels are useful for sorting and harvesting; however they are unable to capture fruit development stages, which limits their use for predicting yield or harvest time. This limitation can be addressed using phenology-based labelling that organizes plants development by biological stages (e.g., BBCH scale). This approach provides necessary information for monitoring crop growth. Phenology-labelled datasets are difficult to build because they require repeated observations and expert annotation, and they are still limited in size and availability [43–45]. Furthermore, differences in growth patterns across crops and grow environments necessitate the development of crop-specific phenology-based datasets.

Table 1 - An overview of the fruit detection datasets in the literature

| Fruit | Dataset Size | Image Size | Farm Type | Availability | Phenology-based |
|---|---|---|---|---|---|
| Apple | 300 | 640×640 | Conventional | Private | No [46] |
| Apple | 1386 | 600×400 | Conventional | Private | No [47] |
| Apple | 267 | 416×416 | Conventional | Private | No [48] |
| Apple | 1200 | 416×416 | Conventional | Private | No [49] |
| Apple | 2298 | 1280×960 | Conventional | Public | No [50] |
| Apple | 1000 | 1280×720 | Conventional | Public | No [41] |
| Apple | 285 | 640×480 | Greenhouse | Private | No [51] |
| Citrus | 4855 | 1920×1080 | Conventional | Public | No [52] |
| Citrus | 579 | 2448×3264 | Conventional | Public | No [53] |
| Grape | 300 | 1365×2048 | Conventional | Public | No [36] |
| Grape | 961 | 300×450 | Conventional | Private | No [29] |
| Mango | 1100 | 800×600 | Conventional | Private | No [30] |
| Raspberry | 2039 | 1773×1773 | Conventional | Public | No [42] |
| Strawberry | 2400 | N.A. | Conventional | Private | No [32] |
| Strawberry | 177 | 1280×720 | Greenhouse | Public | No [35] |
| Tomato | 996 | 512×512 | Greenhouse | Private | No [31] |
| Tomato | 318 | 504×377 | Greenhouse | Private | No [54] |
| Tomato | 250 | 2000×2000 | Greenhouse | Public | No [37] |
| **Raspberry PhenoSet** | **1853** | **5184×3456** | **Vertical Farm** | **Public** | Yes |

To address the challenges associated with plant yield forecasting, harvest time estimation, growth monitoring, and facilitate supply chain management, this study develops a phenology-based dataset suitable for both object detection and semantic segmentation, which enables raspberry yield prediction and harvest time estimation 28 to 32 days in advance of fruit harvest. In addition, the study defines and benchmarks the challenging task of discriminating between subtle developmental transitions using state-of-the-art models through extensive comparative evaluation and depicts the optimal model and configuration for real-time inference that can be integrated into camera-based monitoring systems or robot-based mobile platforms. An overview of the study is illustrated in Fig. 1 and the main contributions of this research can be summarized as:

- Phenology Framework: A phenological timeline was established for three raspberry cultivars (Polana, Prelude, and Joan J) based on the BBCH scale, defining seven developmental stages of raspberry fruit growth.
- Dataset Curation and Public Release: A large, high-quality raspberry image dataset was curated and made publicly available. It contains 1,853 images with 6,907 manually annotated instances across seven phenological stages, with both bounding-box annotations (for object detection) and mask annotations (for segmentation).
- Comprehensive Benchmarking: A systematic comparative study was conducted across state-of-the-art detection and segmentation networks, resulting in a comprehensive benchmark for phenology-stage recognition in raspberries.
- Ready-to-Deploy Model: A lightweight model was developed by fine-tuning YOLOv8 for raspberry phenology-stage detection. The model achieved 91.2% detection accuracy and supports real-time inference on consumer-grade hardware. The trained models have also been made publicly available at the project website (https://sites.google.com/view/raspberry-phenoset)

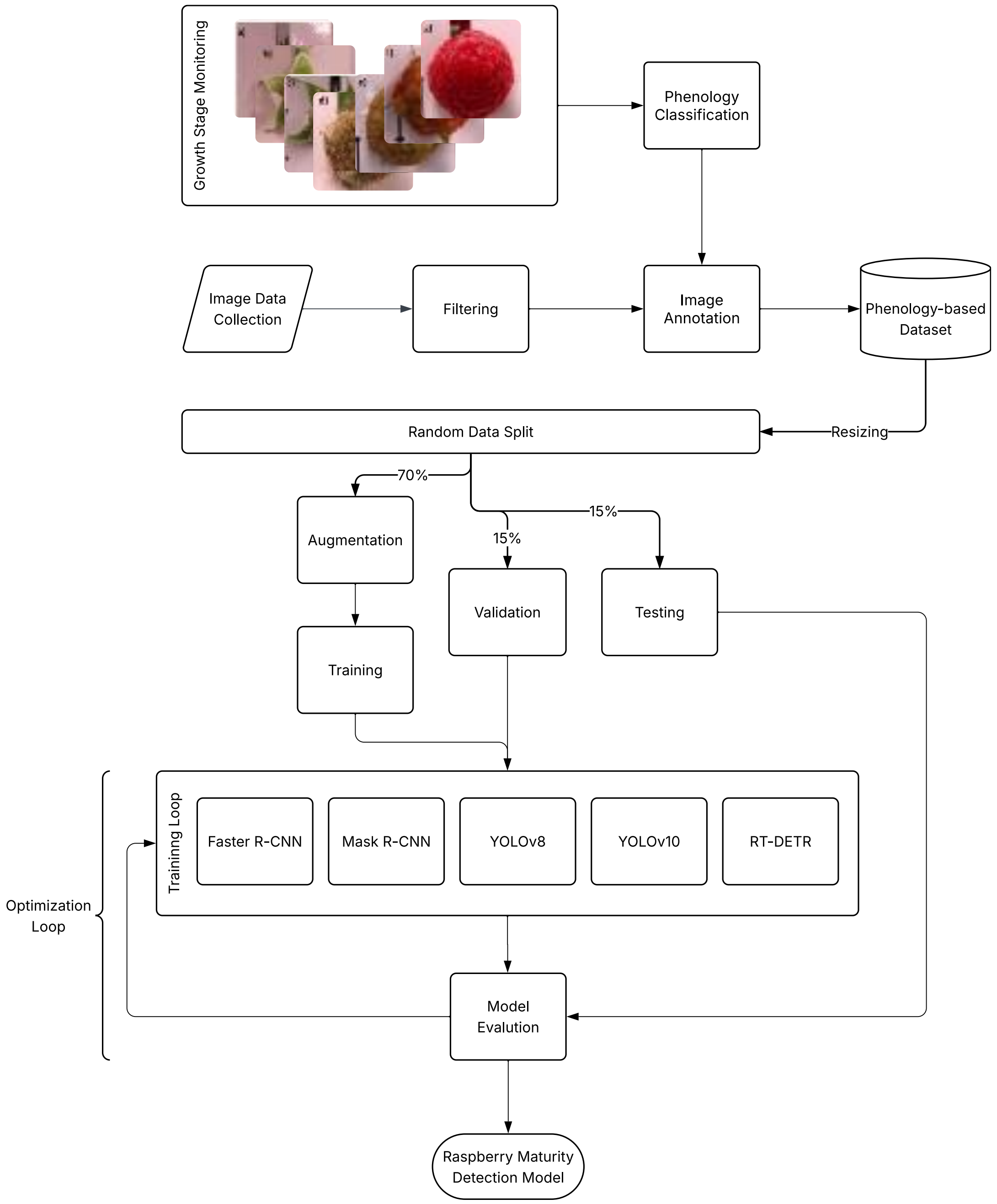

Fig. 1 - Flowchart of the research, outlining the key steps for developing a deep learning raspberry maturity detection model to assist farmers in monitoring and supply chain management tasks.

# 2 Materials and Methods

## 2.1 Study Organisms

The primocane-fruiting red raspberry (*Rubus idaeus* L.) cultivars 'Joan J' and 'Polana' and floricane-fruiting 'Prelude' were grown in our system from bare root plants (Lareault Nursery, Lavaltrie, QC, Canada). 'Joan J' is a product of crossing 'Joan Squire' and 'Terri-Louise' in 1993, and is characterized by medium red, broad conical berries of medium firmness, supported by plants with medium green foliage [55]. 'Prelude' is an early-maturing floricane-fruiting variety that typically reaches 50% maturity approximately 4 months after planting [56, 57]. It has a high percentage of marketable fruit yield (79%) and matures earlier than other varieties, such as 'Canby' and 'Killarney' [56–58]. 'Polana,' an early primocane-fruiting variety, matures in about five months and is known for its high productivity and flavour quality [59–62]. These cultivars were selected for their early fruiting period, relatively large fruit size and flavour, high leaf chlorophyll content, productivity, and growth habit. 'Joan J', 'Polana' and 'Prelude' are early fruiting cultivars, suggesting it would be efficient in a vertical farm where it is important to minimize the time to fruiting [56, 58, 59, 63]

## 2.2 Growing Conditions

The experiment was conducted in 2023-2024 in the Botany Laboratory (43.66° N, -79.38° W) located at the Centre of Urban Innovation, Toronto Metropolitan University, Toronto, Ontario, Canada. 'Polana' and 'Prelude' rootstocks were planted individually in a 3:7 ratio of perlite and vermiculite (Holiday, Mississauga, Canada), whereas 'Joan J' rootstocks were planted in hemp fibre material. They were planted in 2-gallon pots (GrowerBasics, Mississauga, Canada) on November 23 and 24, 2023. Rootstocks were grown at 50% humidity under increasing light intensities for the first week until they received a full light recipe of 195 μmol $m^{-2}$ $s^{-1}$ (90 Cool White + 105 Dark Red) with a 16:8-hour day: night cycle. Throughout the experiment, plants were grown in multiple indoor growth chambers with the dimensions of 254×67.3×102 $cm^3$ (L×W×H) that housed 60 plants each. Once planted, plants were irrigated and fertilized using drip irrigation to maintain soil moisture levels at 15-35% (approximately 500 mL of water every two days). A combination of Magnific (11-0-0 + 9.3% Mg; ICL Premium Fertilizers NA, St. Louis, USA), Calcinit (15.5-0-0 + 19% Ca; Yara North America, Tampa, USA), NK fertilizer (13.5-0-46.2; Haifa Group, Orlando, USA), and MKP fertilizer (0-52-34; Haifa Group, Orlando, USA) crystal fertilizers was used, following the FloraSeries® "Medium Feed" Feed Chart recipe (General Hydroponics Inc., Santa Rosa, United States). The pH of nutrient solutions was checked three times per week and adjusted to 6.0 [64]. Plants were grown under controlled conditions of 21°C ± 3°C and 50% relative humidity [65]. Plants were illuminated under deep red and cool white LED lamps (190 μmol $m^{-2}$ $s^{-1}$), following a 14-day light ramp-up schedule for acclimatization, as described by Bazangeya et al. [66].

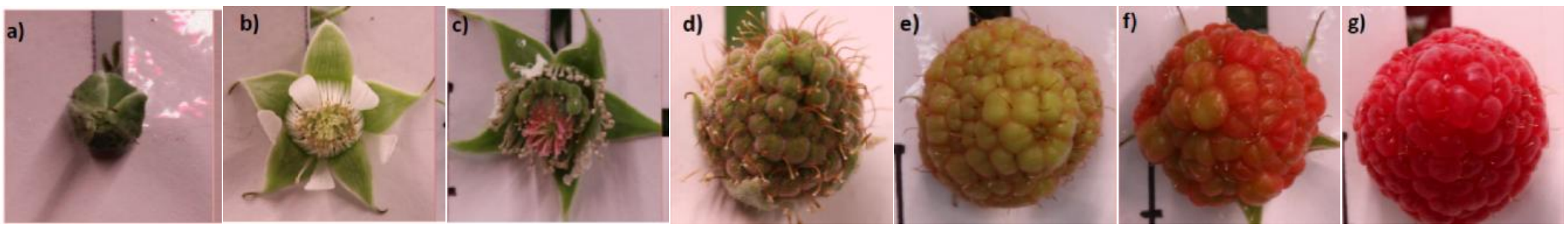

Fig. 2 - Phenology reference of raspberry development stages used for annotating the images. The stages are labelled a-g, representing the seven development stages. **a)** Bud **b)** Open Flower **c)** Fruit Initiation **d)** Green Fruit **e)** Growing Fruit (yellow colour) **f)** Semi-mature (pink) fruit **g)** Mature (red) fruit.

## 2.3 Phenological Observation Methods

A detailed study of the phenological stages of raspberries was carried out by taking photographs of the plants throughout the growth period. On January 26, 2023, five flower buds were haphazardly selected on randomly selected 'Joan J' raspberry plants and tagged with a paper tag. On June 28th, 2023, one bud from nine randomly selected 'Polana' and nine randomly selected 'Prelude' raspberry plants was tagged with a light string. The buds were haphazardly selected from the top half of the plant. The progression from a small bud to ripe fruit was documented through daily photo capture until the fruit reached full maturity. The duration (in days) of each phenological stage on each plant was documented through visual observations and photo capture according to the BBCH classification of Schmid et al. [67] and adopted from Rotili et al. [68]. The study resulted in seven distinct stages, with each stage displaying the following distinctive characteristics:

- Stage A: floral buds (BBCH #50-59).
- Stage B: open flower with white petals (BBCH #60-65).
- Stage C: green fruit initiation with very small drupelets. Petals may or may not be attached (BBCH #69).
- Stage D: green fruit with drupelet growth (BBCH #70-79).
- Stage E: growing fruit with colour changing from green to yellow (BBCH #81).
- Stage F: semi-mature pink fruits (BBCH #85).
- Stage G: mature red fruit (BBCH #89).

Further, 80 buds from different cultivars were followed to document their date of first expression of each of the phenological stages. The growth stages were consistent between the cultivars (Fig. 2).

## 2.4 Data Acquisition

In the next step of this study, the same three cultivars were grown in our vertical farm facility at the Centre for Urban Innovation at Toronto Metropolitan University from January to March 2024, and images were captured as the plants grew. Since the visual differences between cultivars were negligible, they were merged to form a larger and more robust dataset called Raspberry PhenoSet. A total of 2200 images were captured for further processing. The images were taken with a handheld Canon EOS Rebel T3i at a 5184×3456 resolution, which has a variable focal length of 18-55 mm and an electronically controlled focal-plane shutter. The camera was set to Scene Intelligent Automatic Mode, which determined the appropriate aperture, shutter speed, and ISO to adjust exposure. The metadata of each image is saved and accessible in its properties; camera calibration results are also provided with the dataset to assist in the recreation of the research. During the photo sessions, overhead lights were the main source of lighting, and no additional lighting was used to maintain the illumination intact.

The vertical farming facility uses artificial lights instead of sunlight, providing constant light intensity during the day to enhance plant growth. Nonetheless, due to raspberry plants growing and fruiting in various directions, the dataset includes photos of fruits with different levels of illumination. Also, due to the presence of air nozzles, irrigation systems, and other subsystems of the vertical farm, many images are occluded with unwanted obstacles. The photos were taken at various distances, ranging from 20 to 100 cm from the target objects, resulting in various sizes for the objects of interest. Moreover, since the camera was set on auto-focus, some objects of interest became blurry. All these factors contributed to having a realistic and diverse dataset that represents the vision challenges that a vision system or a robot may face while operating in vertical farming environments.

## 2.5 Data Processing

### 2.5.1 Phenology Analysis:

To provide an initial overview of phenological progression and assess cultivar-level variability, the mean, standard deviation, and range of durations (in days) for each developmental stage were calculated, ranging from bud emergence (Stage A) through to mature fruit (Stage G), across the three raspberry cultivars (‘Polana’, ‘Prelude’, and ‘Joan J’).

### 2.5.2 Classification Criteria

After discarding the images with a high degree of blurriness or other corruption, 1853 images were selected for annotation. Using phenology findings, all visible and recognizable instances of buds, flowers, and fruits in all photographs were annotated; a sample image from the dataset is presented in Fig. 3. These instances were categorized into one of the seven stages found in the phenology study (refer to Fig. 2). The V7Labs annotation platform [69] was utilized for labelling, and objects were annotated using the polygon format to ensure that the details were captured properly. Nonetheless, annotations are provided both in the bounding box and mask format to make the dataset suitable for both object detection and segmentation algorithms.

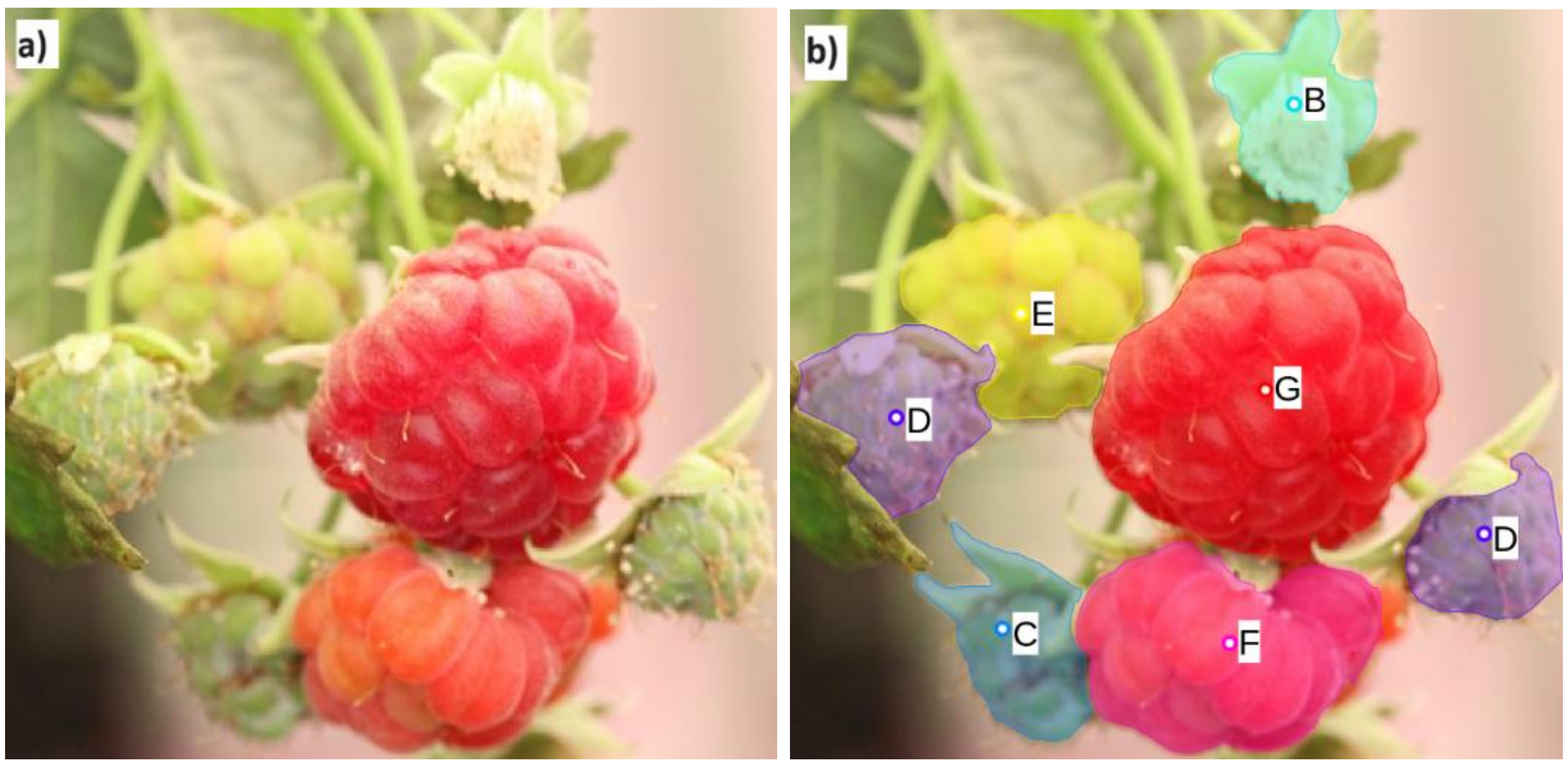

Fig. 3 - A sample image from the Raspberry PhenoSet, taken at the vertical farming facility at Toronto Metropolitan University. **a)** Original image with no annotations. **b)** Mask annotations of all instances present in the image; Labels A-G correspond to the seven phenology stages of raspberries.

### 2.5.3 Statistical Report

Due to the continuous development of new buds, flowers and fruits throughout the documentation period, often there is more than one phenology stage present in the images. A statistical report of the dataset is provided in Table 2. The *Total images* column reports the number of images captured while the majority of the plants were at a specific stage, whereas the *Total annotations* column reports the total number of instances of each stage in all images. The dataset consists of 1853 images and contains 6907 annotations. The first three stages have a higher number of annotations due to the continuous emergence of new buds throughout the three-month growth period. Since the images were captured at various distances from the plants, the Raspberry PhenoSet includes annotations at various sizes, starting from 20 pixels and reaching more than 1500 pixels. This allows the networks to learn the target objects' details and shape factors properly. A histogram of annotations is illustrated in Fig. 4 to provide a more comprehensive report.

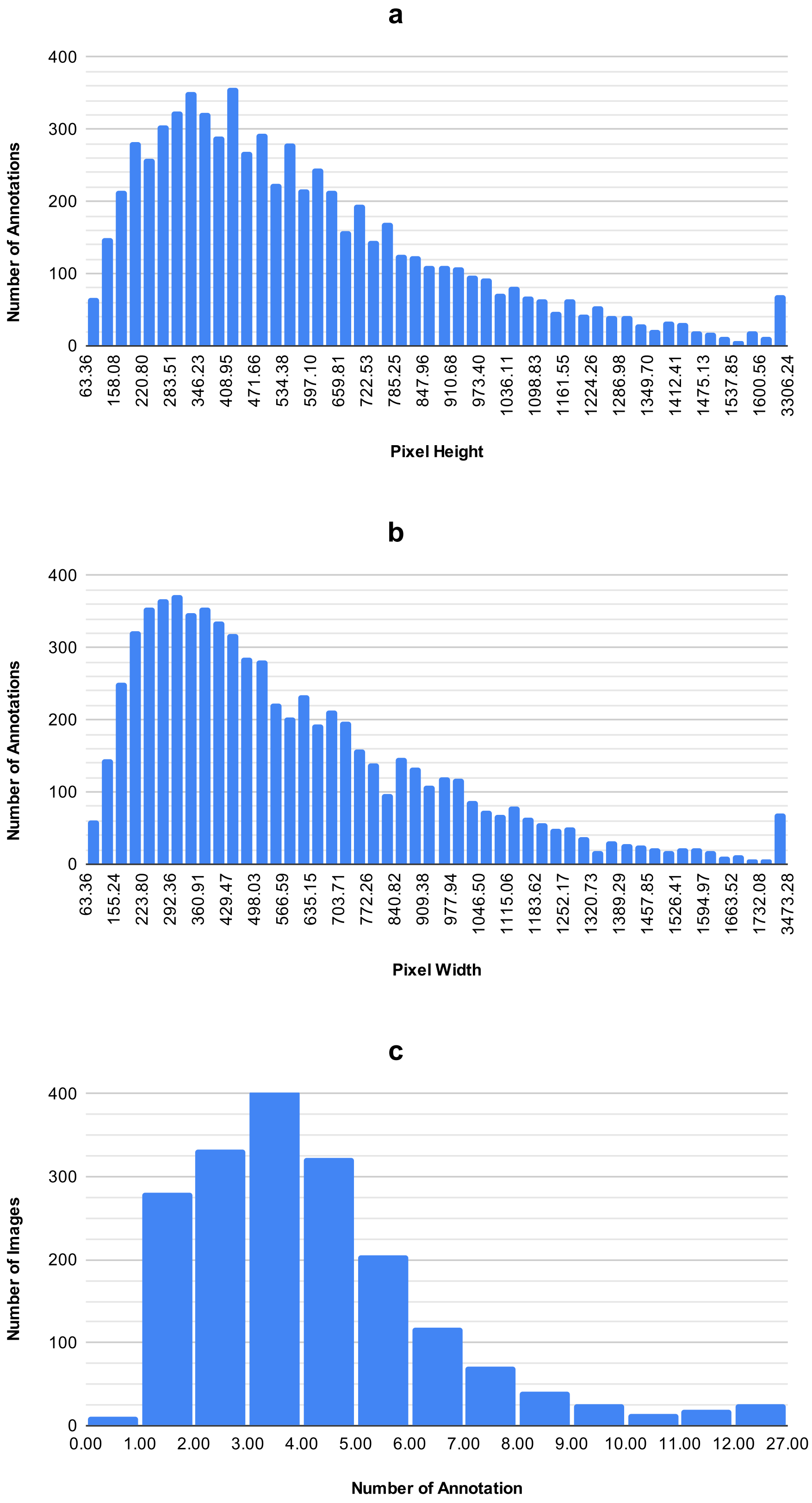

Fig. 4 - Histogram of Annotations in Raspberry PhenoSet with 1% outlier. **a)** Number of Annotations vs. Pixel Heights **b)** Number of Annotations vs. Pixel Widths **c)** Number of Images vs. Number of Annotations per Image

Table 2 - Specification of raspberry Phenoset annotations

| Phenology Stage | Total Images | Total Annotations | Mean BB Width [px] | Mean BB Height [px] | Average Annotated Area [px$^2$] | Average Image Area [%] |
|---|---|---|---|---|---|---|
| A | 187 | 1322 | 315.03 | 352.87 | 135075.68 | 0.75 |
| B | 452 | 1428 | 688.25 | 668.81 | 570569.68 | 3.18 |
| C | 373 | 1424 | 549.07 | 562.20 | 404791.96 | 2.26 |
| D | 260 | 1267 | 716.96 | 689.11 | 602083.71 | 3.36 |
| E | 218 | 509 | 661.56 | 646.58 | 544447.26 | 3.04 |
| F | 148 | 386 | 726.56 | 700.80 | 652934.44 | 3.64 |
| G | 215 | 571 | 508.94 | 496.18 | 325503.06 | 1.82 |
| **Total** | **1853** | **6907** | **595.20** | **588.10** | **462200.83** | **2.58** |

#### 2.5.4 Augmentation and resizing

Since all the training was done locally using a consumer-grade computer (more details available on Sec. 2.8), images were converted to 900×600 for training purposes. This conversion prevented the system from running out of memory while maintaining the same aspect ratio as the original images. Thus, the images are provided in two sizes: a smaller size suitable for widely used GPUs and a larger size suitable for systems with high graphical processing power and future hardware upgrades.

Random augmentations such as translation, rotation, flip, brightness adjustment, shear deformation, and Gaussian noise were implemented on the dataset, with a probability of 10%, using the OpenCV library in Python, with the aim of increasing the generalizability of models and preventing them from overfitting. Nonetheless, the training was carried out using both augmented and original datasets to study the effect of the augmentation.

### 2.6 Deep learning models

State-of-the-art object detection models can be categorized as either one-stage or two-stage networks. One-stage detectors perform object detection in a single pass through the network, predicting bounding boxes and class probabilities directly from the input image. One-stage detectors are typically faster and more suitable for real-time applications because they process images in a single forward pass. [70]. Two-stage detectors use a more complex approach, first generating region proposals and then refining these proposals to detect objects. Two-stage detectors generally achieve higher accuracy and better performance in terms of localization and classification, particularly for small and densely packed objects [71]. To establish an optimal maturity detection model, a systematic and comparative study was conducted, and object detection architectures were optimized. For this purpose, a combination of top object detection and segmentation models was chosen based on their characteristics and their performance was evaluated on the Raspberry PhenoSet to provide a comprehensive benchmark. Moreover, an optimization loop was formed to obtain the best performance of object detection architectures on raspberry maturity detection.

#### 2.6.1 One-stage networks

This field is dominated by different versions of the YOLO network [72]. In this paper, YOLOv10, YOLOv8, and RT-DETR were chosen as the one-stage object detection models. YOLOv10 was chosen due to its real-time capabilities [73] and YOLOv8 was chosen due to having a greater number of parameters [74], and RT-DETR was chosen since it uses a transformer-based architecture, which is different from YOLOs that are based on convolutional neural networks [75].

#### 2.6.2 Two-stage networks

Faster R-CNN and Mask R-CNN are the top two-stage networks in the literature, and they are commonly used with ResNet-50 and ResNet-101 backbones [71]. Faster R-CNN was primarily designed for object detection, while Mask R-CNN extends Faster R-CNN to add instance segmentation, enabling it not only to detect objects but also to generate a precise segmentation mask for each detected object [76, 77]. Although ResNet-101 is a larger network and is expected to perform better, both backbones have been tested to provide a comparison.

### 2.7 Evaluation metrics

To evaluate and compare the performance of different models, the images in each class were randomly split into 70%, 15%, and 15% portions for training, validation, and testing purposes, respectively and the following metrics were used to quantify the models' performance.

#### 2.7.1 Precision (P):

A performance metric calculated by dividing the number of true positives by the sum of the true positives and false positives. Predictions with an intersection over union (IoU) values of 0.5 or greater were accepted as true positives, and the rest were counted as false positives.

$$P = \frac{TP}{TP + FP} \quad (1)$$

#### 2.7.2 Recall (R):

Another performance metric indicating how often the model detects positive instances correctly and is calculated by dividing the number of true positives (TP) by the sum of true positives and false negatives (FN).

$$R = \frac{TP}{TP + FN} \quad (2)$$

#### 2.7.3 Average Precision (AP):

A more comprehensive performance metric used primarily to evaluate models in object detection and other classification tasks. AP summarizes the precision-recall curve by calculating the area under the curve. AP provides a single value representing the model's ability to balance precision and recall across different thresholds.

$$AP = \int_0^1 P(R)dR \quad (3)$$

#### 2.7.4 Mean Average Precision (mAP):

This metric is the mean of the AP values for multiple classes and is useful in multi-class object detection to provide a comprehensive evaluation of the model's performance across all classes. mAP is calculated as:

$$mAP = \frac{1}{N}\Sigma_{i=1}^{N} AP_i \quad (4)$$

where N is the number of classes and $AP_i$ is the average precision for class $i$.

2.7.5 F1-score:

This metric is used to evaluate the performance of a classification model, balancing both precision and recall. It is the harmonic mean of precision and recall, providing a single score that accounts for both false positives and false negatives.

$$F_1 = 2 \times \frac{P \times R}{P + R} \tag{5}$$

## 2.8 Experimental setup

All models were trained locally on a computer with a 12th Gen Intel Processor, i7-12700x20, 16 GB of RAM and an Nvidia GeForce RTX-3060 GPU with 12 GB of memory running on Ubuntu 22.04.5 LTS 64-bit. The networks' hyperparameters were adjusted to make the training feasible, due to memory limits, and to improve the performance of the networks. Recommended values for learning rate, weight decay, and Adam W optimizer were used for the training of the models. All models used in this study have a variable learning rate, i.e. they take an initial learning rate but use the calculated learning rate obtained from their optimization algorithms (hyper-parameters are summarized in Table 3).

Table 3 - Hyperparameters of implemented models

| Model | Backbone or Scale | Batch Size | Learning Rate | Weight Decay | Epochs |
|---|---|---|---|---|---|
| Faster R-CNN | ResNet-50 | 8 | 0.001 | 0.0001 | 150 |
| Faster R-CNN | ResNet-101 | 6 | 0.001 | 0.0001 | 150 |
| Mask R-CNN | ResNet-50 | 8 | 0.001 | 0.0001 | 150 |
| Mask R-CNN | ResNet-101 | 4 | 0.001 | 0.0001 | 150 |
| YOLOv8 | yolov8-m/l | 16 | 0.001 | 0.0005 | 100 |
| YOLOv8 | yolov8-x | 12 | 0.001 | 0.0005 | 100 |
| YOLOv10 | yolov10-l | 12 | 0.001 | 0.0005 | 100 |
| YOLOv10 | yolov10-x | 9 | 0.001 | 0.0005 | 100 |
| RT-DETR | rtdetr-l/x | 12 | 0.001 | 0.0005 | 100 |

# 3 Results and Discussion

Raspberries are a high-value, rapidly perishable crop with narrow harvest windows (2–5 days post-maturity), creating logistical pressure for growers and distributors to predict harvest timing accurately. Yet phenological tracking remains largely manual, prone to error, and difficult to scale, especially in high-density environments like vertical farms. In this study, Raspberry PhenoSet was developed to enable automated, fine-grained detection of phenology stages using state-of-the-art object detection models. Moreover, deep learning models were trained and optimized to establish the raspberry maturity detection model. To enhance the performance of object detection architectures, transfer learning was leveraged by using pre-trained weights from each network's official GitHub repository. The following results demonstrate the temporal progression of three commercial raspberry cultivars and evaluate model performance in detecting biologically meaningful stages that correlate with days-to-harvest, a prerequisite for supply chain-aligned automation.

### 3.1 Phenological progression across cultivars

Three raspberry cultivars, ‘Joan J’, ‘Polana’, and ‘Prelude’, exhibited distinct phenological patterns under controlled vertical farm conditions. Although the phenology stages were consistent, the timelines were different for each cultivar. For instance, ‘Joan J’ exhibited the shortest time to maturity (26–29 days), with rapid progression through flowering (2.0 days ± 1.6 SD) and early fruit development stages. This makes it ideal for tightly scheduled harvest cycles. By contrast, ‘Polana’ and ‘Prelude’ required 32–38 days to reach maturity, offering more prolonged harvest periods, which can buffer risk or align with extended market availability. Notably, ‘Polana’ showed a long and variable Stage D (13.3 days ± 12.1 SD), which can introduce uncertainty into forecasting models unless regularly monitored (Table 4).

Table 4 - Phenology-based correlation of growth stage and harvest time for different raspberry cultivars

| | Days Left to Harvest for Cultivars | | |
|---|---|---|---|
| **Phenology Stage** | **Prelude** | **Polana** | **Joan J** |
| A | 31.9 ± 0.35 | 33.3 ± 0.71 | 28.0 ± 0.41 |
| B | 28.5 ± 2.37 | 31.1 ± 1.94 | 25.6 ± 1.64 |
| C | 25.1 ± 1.5 | 27.5 ± 2.55 | 23.6 ± 5.12 |
| D | 13.6 ± 7.76 | 16.5 ± 12.12 | 13.5 ± 3.13 |
| E | 5.0 ± 4.66 | 3.2 ± 3.74 | 3.3 ± 2.06 |
| F | 2.8 ± 0.0 | 0.5 ± 3.01 | n/a |
| G | 0 | 0 | 0 |

These cultivar-specific growth rates, when annotated across thousands of images, form the empirical foundation for training models that map visual cues to temporal harvest projections. This enables predictive harvest timing, a critical capability for minimizing food loss, negotiating delivery contracts, and planning distribution logistics in perishable fruit supply chains.

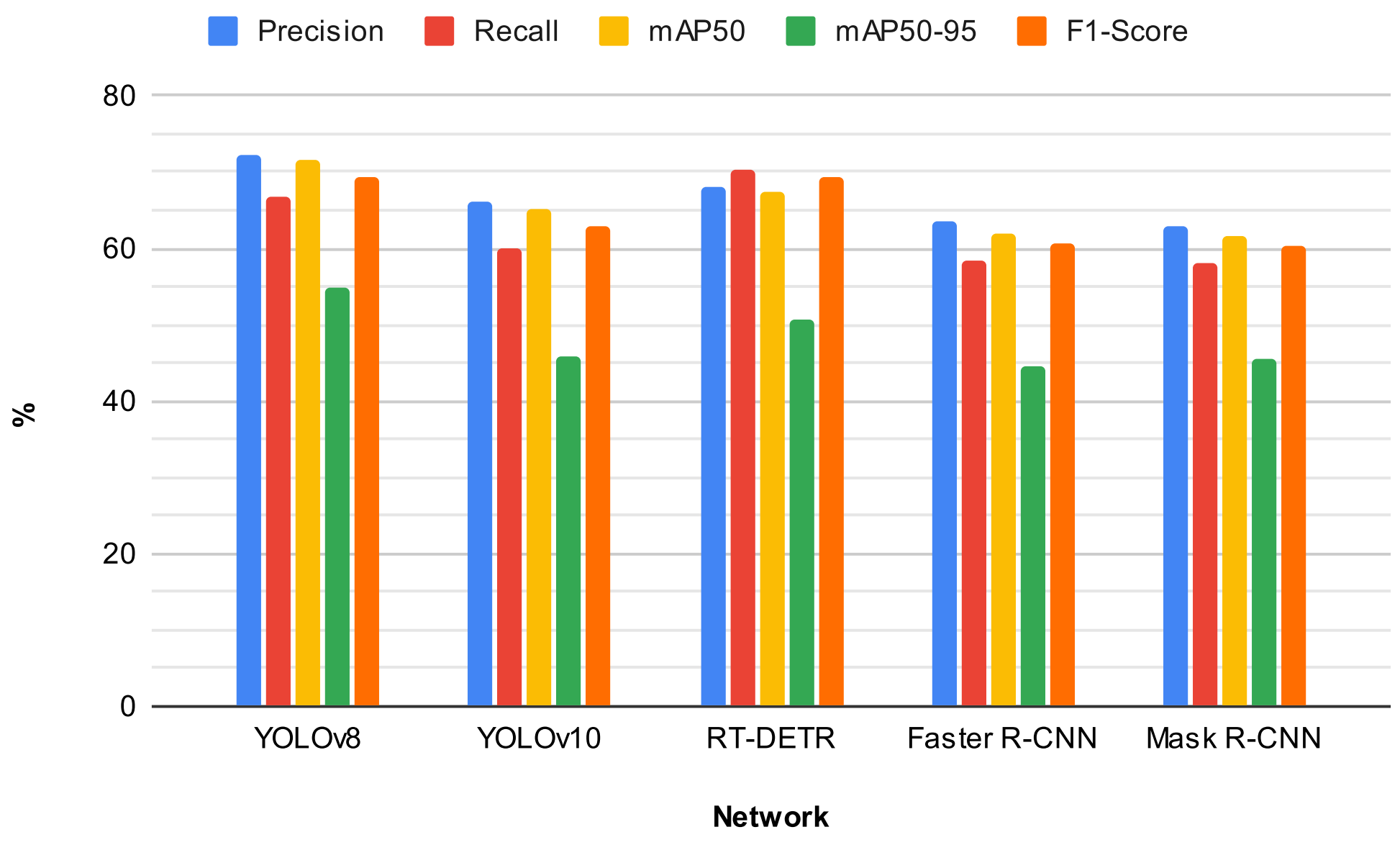

Fig. 5 - Best Performance of Different Networks on the Raspberry PhenoSet, Visualizing the Suitability of YOLOv8 and RT-DETR for Fruit Detection Applications

## 3.2 Performance of one-stage models

The performance of three state-of-the-art one-stage object detection architectures, YOLOv8, YOLOv10, and RT-DETR, were benchmarked on the Raspberry PhenoSet dataset. These models were selected based on the trade-offs between detection accuracy and computational efficiency by prioritizing edge computing deployment. Each model was trained and evaluated at multiple scales to assess scalability and performance under varying configurations. Furthermore, model parameters were optimized to achieve the highest performance metrics.

Among the tested configurations, YOLOv8-x achieved the highest detection performance, with a precision of 0.721, recall of 0.668, and mAP50 of 0.717 (Table 5, and Fig. 5). It also achieved strong stage-level performance, with an F1 score of 0.810 for floral buds (Stage A) and 0.739 for mature fruit (Stage G) (Table 6). These stages are particularly important for forecasting and scheduling in raspberry production, as Stage A is used to initiate yield estimation models, and Stage G represents the narrow harvest window critical to reducing spoilage and economic loss.

Table 5 - Performance of different networks on the raspberry PhenoSet

| Network | Backbone/Scale | Precision | Recall | mAP50 | mAP50-95 | F1 |
|---|---|---|---|---|---|---|
| YOLOv8 | yolov8-m | 0.683 | 0.661 | 0.698 | 0.542 | 0.672 |
| YOLOv8 | yolov8-l | 0.694 | 0.674 | 0.700 | 0.544 | 0.684 |
| YOLOv8 | yolov8-x | **0.721** | 0.668 | **0.717** | **0.550** | **0.693** |
| YOLOv10 | yolov10-l | 0.661 | 0.599 | 0.652 | 0.460 | 0.628 |
| YOLOv10 | yolov10-x | 0.602 | 0.653 | 0.647 | 0.457 | 0.626 |
| RT-DETR | rtdetr-l | 0.681 | 0.700 | 0.659 | 0.512 | 0.690 |
| RT-DETR | rtdetr-x | 0.682 | **0.704** | 0.674 | 0.506 | **0.693** |
| Faster R-CNN | ResNet-50 | 0.634 | 0.583 | 0.620 | 0.446 | 0.607 |
| Faster R-CNN | ResNet-101 | 0.594 | 0.577 | 0.591 | 0.430 | 0.585 |
| Mask R-CNN | ResNet-50 | 0.628 | 0.580 | 0.617 | 0.454 | 0.603 |
| Mask R-CNN | ResNet-101 | 0.583 | 0.575 | 0.571 | 0.442 | 0.579 |

RT-DETR-x demonstrated the highest overall F1 score (0.693) across classes but required greater computational resources due to its transformer-based architecture (Table 5). YOLOv10, optimized for real-time inference, showed lower accuracy, a limitation attributed to reduced batch sizes (Table 3) necessitated by higher GPU memory demands at larger scales (Table 5).

Table 6 - Performance of YOLOv8-x on the Raspberry PhenoSet

| Class | Precision | Recall | AP50 | mAP50-95 | F1 |
|---|---|---|---|---|---|
| A | 0.802 | 0.819 | 0.855 | 0.679 | 0.810 |
| B | 0.773 | 0.635 | 0.729 | 0.493 | 0.697 |
| C | 0.621 | 0.534 | 0.592 | 0.367 | 0.574 |
| D | 0.641 | 0.652 | 0.677 | 0.482 | 0.646 |
| E | 0.731 | 0.569 | 0.640 | 0.506 | 0.640 |
| F | 0.849 | 0.577 | 0.678 | 0.599 | 0.687 |
| G | 0.632 | 0.891 | 0.850 | 0.712 | 0.739 |
| **Mean** | 0.721 | 0.668 | 0.717 | 0.548 | 0.693 |

Overall, YOLOv8 exhibited the most favourable balance of detection accuracy, model efficiency, and adaptability to raspberry phenology stages, as illustrated in Fig. 6. Its ability to reliably distinguish between visually similar stages, such as green fruit swelling (Stage D) and early pigment development (Stage E), is essential for time-sensitive applications such as harvest prediction and robotic picking. However, performance limitations remain.

Analysis of the confusion matrix (Fig. 7) indicates that stages with subtle visual differences are frequently misclassified, and false negatives are most common in early stages such as buds and flowers, likely due to visual overlap with the background (e.g., leaves or white shelving).

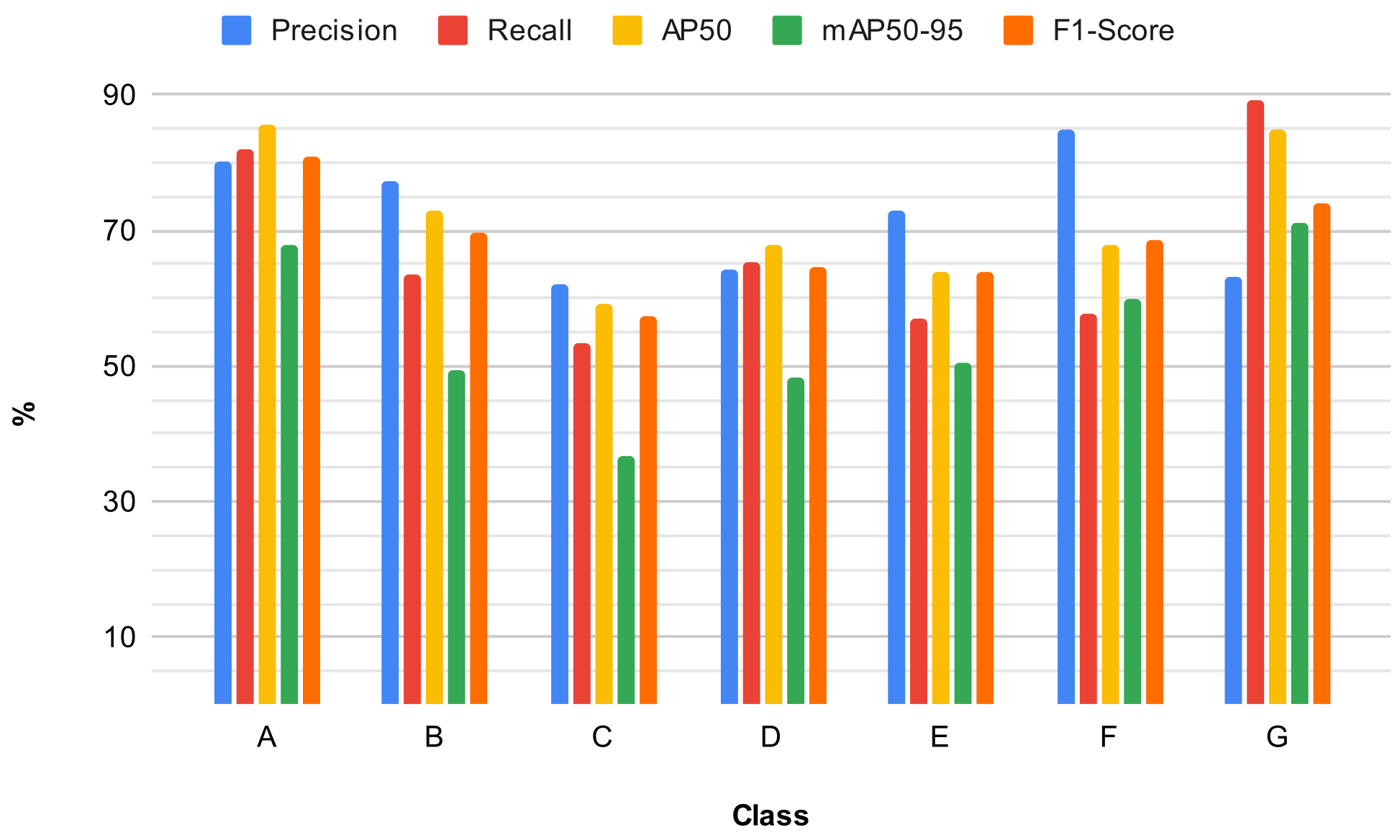

Fig. 6 - Performance of YOLOv8-x on Each Class of the Raspberry PhenoSet, Showing the Model's Ability to Distinguish Each Phenology Stage

### 3.3 Performance of two-stage models

To study the applicability of two-stage object detection architectures for phenology-stage classification, Faster R-CNN and Mask R-CNN were trained using ResNet-50 and ResNet-101 backbones. These models are commonly employed for high-precision object detection tasks, particularly in domains involving small, densely packed objects. However, in testing the two-stage models, both architectures exhibited lower overall performance with Raspberry PhenoSet compared to one-stage models such as YOLOv8 and RT-DETR (Table 5).

Models using the deeper ResNet-101 backbone consistently underperformed relative to the ResNet-50 counterparts, a result attributed to increased computational overhead that necessitated reduced batch sizes during training (Table 3). This resource constraint limited the ability of the deeper networks to converge effectively. This underscores the importance of light-weight architecture design that can run efficiently on consumer-grade computers. Across all configurations, the two-stage models produced lower precision, recall, and F1 scores than YOLOv8-x and RT-DETR-x. This trend was particularly evident in phenological stages with subtle visual differences, such as fruit swelling (Stage D) and early colour development (Stage E), where fine-grained classification accuracy is essential.

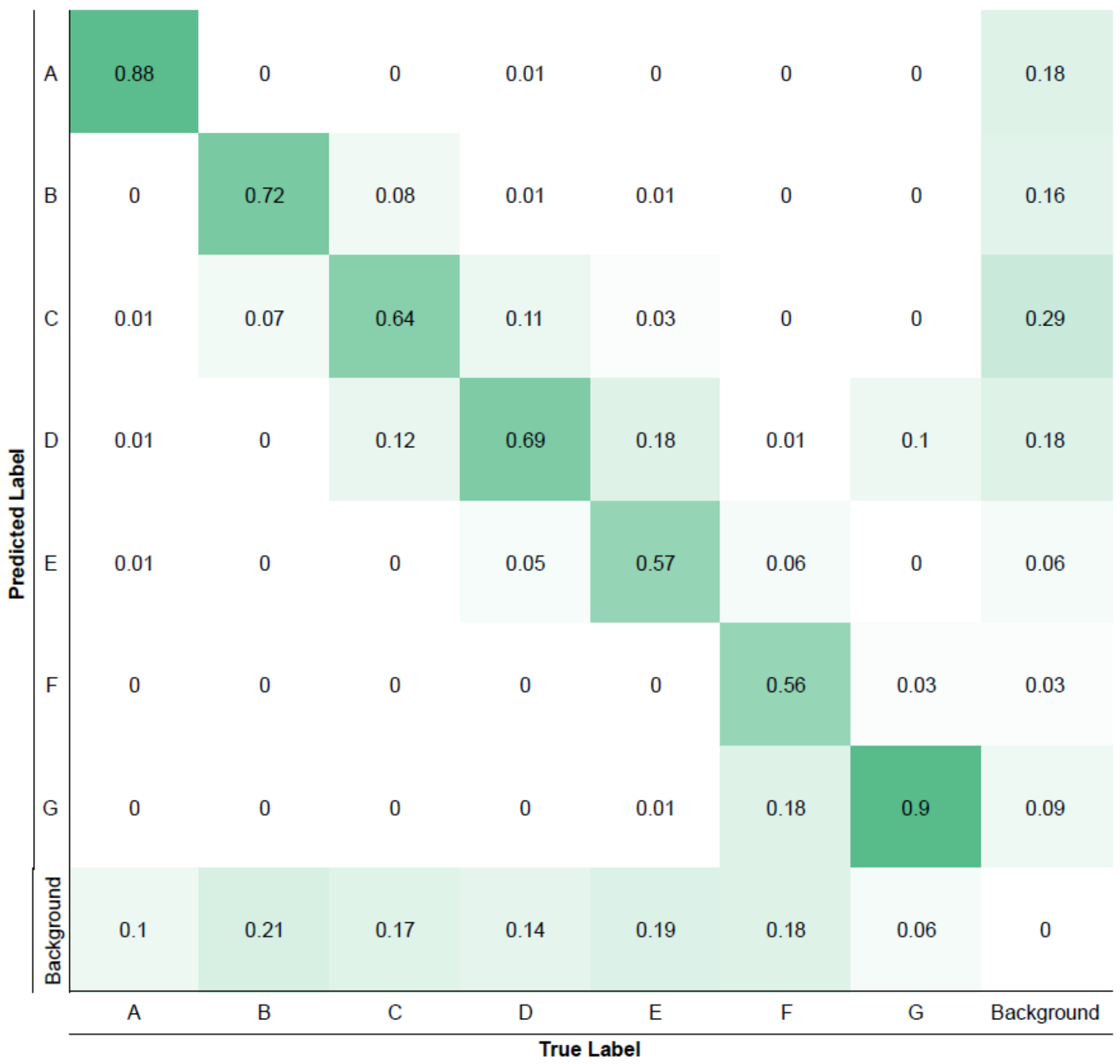

Fig. 7 - Normalized Confusion Matrix of YOLOv8-x. The values are normalized by the total number of true instances for each class (column normalization), representing a recall-oriented confusion matrix

### 3.4 Phenology-Based Dataset Insights and Limitations

The results of this study highlight both the potential and the challenges associated with phenology-based labelling in agricultural computer vision. By structuring the dataset around several BBCH-aligned developmental stages, the Raspberry PhenoSet enables the model to learn biologically meaningful transitions that can be directly linked to time-to-harvest predictions, addressing a key limitation of conventional fruit detection datasets that rely on coarse maturity labels. At the same time, our findings reflect several limitations commonly reported in phenology-based approaches. For example, the confusion observed between adjacent developmental stages (e.g., Stage D and Stage E) underscores the difficulty of distinguishing subtle morphological transitions, particularly when labelled datasets are limited, and stage boundaries are visually ambiguous [45, 78, 79].

Similarly, the variability in stage duration across cultivars (Table 4) highlights the importance of aligning phenological timelines when extending models across genotypes or environments, as misalignment of phenological signals has been shown to reduce model transferability across crops and years [80]. These observations are consistent with broader findings that phenology-based tasks require well-aligned datasets to achieve robust generalization [43, 79, 81]. In this context, the Raspberry PhenoSet contributes toward addressing dataset scarcity in fruit crops by providing a high-resolution, phenology-grounded benchmark that supports both detection and forecasting tasks, while also highlighting the need for standardized labelling frameworks and expanded datasets to further improve model performance.

### 3.5 Propagation of Classification Errors to Harvest Time Prediction

Standard metrics such as precision, recall, and mAP quantify the classification performance of the model; however, they do not directly estimate the remaining days to harvest. To overcome this limitation, we analysed how classification errors propagate to harvest time prediction through the phenological stage-to-time mapping.

Let $y \in \{1, \dots, K\}$ denote the true phenological stage and $\hat{y}$ denote the predicted stage. Each stage is associated with a mean remaining time to harvest $T_j$, obtained from the phenology study. The expected absolute error in harvest time due to classification uncertainty across all classes is defined as:

$$E(|\Delta T|) = \sum_{j=1}^{K} P(y = j) \sum_{i=1}^{K} P(\hat{y} = i \mid y = j) |T_i - T_j| \quad (6)$$

where $K$ is the total number of classes in the dataset, $T_i$ and $T_j$ denote the mean remaining days to harvest for the true stage $j$ and predicted stage $j$, respectively, and $P(\hat{y} = i \mid y = j)$ is obtained from the confusion matrix. A stage-balanced formulation was adopted by setting $P(y = j) = 1/K$ to reflect the true biological distribution of stage occurrence. This avoids bias toward overrepresented stages and ensures that the propagation analysis reflects the full phenological progression rather than dataset composition.

The resulting expected mean absolute error in remaining days to harvest was approximately 1.30 ± 3.71 days for Prelude, 1.44 ± 4.36 days for Polana, and 1.20 ± 3.57 days for Joan J. These results indicate that, despite moderate classification errors, the induced deviation in harvest time estimation remains relatively small. This behaviour can be attributed to the structure of the confusion matrix, where most misclassifications occur between adjacent phenological stages with similar temporal characteristics. Consequently, the model demonstrates robustness in practical deployment, as classification inaccuracies do not translate into large errors in harvest scheduling.

### 3.6 Field Experiments

To evaluate the proposed model's performance under realistic deployment conditions, the vision system was tested on raspberries grown in a newly patent-pending (PCT/CA2025/050227) mobile vertical farm at Eric Palin Hall at Toronto Metropolitan University. To this end, a laptop configured with a 6th Gen Intel Processor, i7-6700HQ, 16 GB of RAM, and an Nvidia GeForce GTX960M with 4 GB of memory running on Ubuntu 22.04.5 LTS 64-bit, was used as a mobile platform.

Two experiments were conducted: 1) real-time object detection and classification, and 2) offline post-processing object detection and classification. For real-time evaluation, a Logitech C922 Pro webcam was used to capture live images of flowering and fruiting raspberry plants. In the offline test, high-resolution images were collected using a Samsung S23 Ultra (12 MP) camera. The fine-tuned YOLOv8-x model processed real-time frames at 31.8-35.7 ms per image (approximately 28-31 FPS), whereas the fine-tuned YOLOv8-m processed frames at 15.2-19.4 ms (videos are available at https://sites.google.com/view/raspberry-phenoset). In the offline experiment, the proposed model required 28.2-32.7 ms per image. The inference results at the vertical farm are presented in Fig. 8, demonstrating the detection of 47 out of 51 instances present in the farm and yielding an accuracy of 92.2%.

These results confirm that the proposed models can support practical deployment in phenological monitoring systems operating at 30-55 frames per second under real-time conditions, depending on model size. This latency range is compatible with automation tasks such as time-sensitive harvest identification and robotic actuation. Moreover, it should be noted that this performance was achieved using a lower-end laptop to demonstrate the feasibility and practicality of the model. Therefore, using current computing capacity would yield higher frames per second inference.

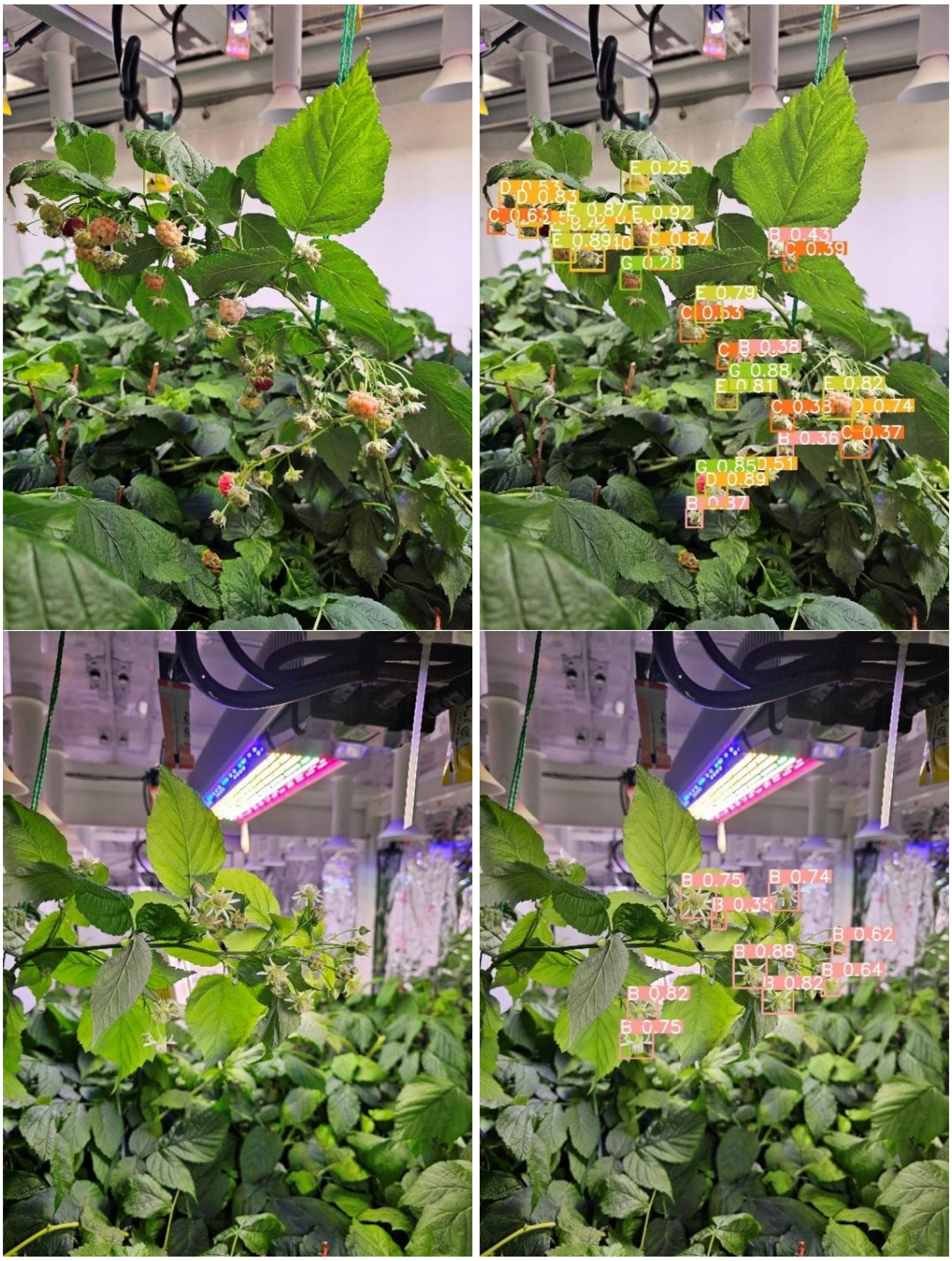

Fig. 8 - Field experiment samples of the trained YOLOv8-x on Raspberry PhenoSet, showing the predicted bounding boxes, labels and confidence scores. Raw images are on the left and the processed images are on the right.

### 3.7 Model generalizability assessment

For further validation and generalization of the machine learning model trained on Raspberry PhenoSet, we have evaluated the model on an independent dataset from a conventional farm [42]. This dataset was gathered at the Institute of Horticulture, Dobele, Latvia (LatHort), under varying natural lighting featuring additional cultivars (such as Glen Ample, Kapriz Bogov, Lina, Octavia, Ruvi, etc.). The model successfully distinguishes multiple fruit development stages as shown in Fig. 9. These results confirm the model's generalizability across different cultivars and environmental conditions including lighting conditions.

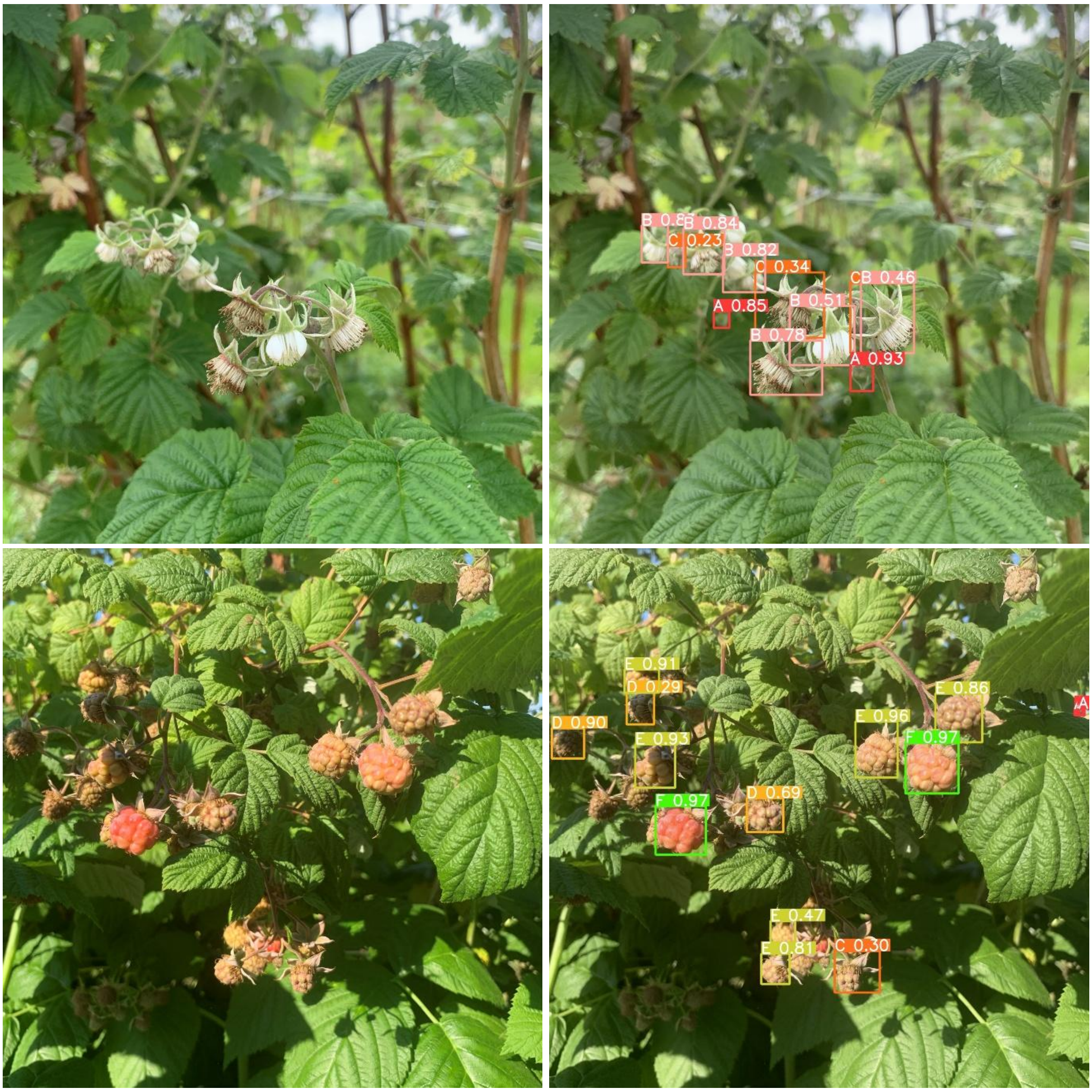

Fig. 9 - Validation of YOLOv8-x trained on Raspberry PhenoSet on an outdoor raspberry dataset [42]. Raw images are on the left, and processed images are on the right, showing the predicted bounding boxes, labels, and confidence scores.

### 3.8 Prospective applications

The global raspberry market was valued at approximately USD 2.14 billion in 2024 and continues to grow due to increasing demand for fresh and processed berries [82]. However, postharvest losses can exceed 20% due to the fruit’s extreme perishability and narrow harvest window [83], making accurate phenological monitoring and timely harvesting economically critical.

Therefore, accurate detection of phenological stages is critical for improving automation in high-value, perishable crops such as raspberries, which typically have a post-harvest shelf life of 2–5 days and are highly sensitive to harvest timing [6]. Even minor deviations in harvest date can lead to reduced fruit quality, increased spoilage, and downstream supply chain inefficiencies [6]. The Raspberry PhenoSet dataset, in combination with high-performing models such as YOLOv8-x, enables several applications aimed at mitigating these risks in sustainable and autonomous controlled-environment agriculture systems.

One important application is dynamic yield forecasting, wherein the abundance and distribution of early-stage fruit structures (e.g., floral buds, flowers, and green fruit) are automatically quantified and used to project future harvest volumes. This enables growers to allocate labour and transformation resources efficiently and to inform retailers of expected delivery schedules, reducing contract penalties and post-harvest waste [83]. As shown in the phenology-to-harvest timelines (Table 4), the known temporal intervals between stages can be used to forecast the likely harvest window from a single phenological snapshot, particularly in cultivars like ‘Joan J’ with predictable development rates.

Second, the models developed here are well-suited to real-time growth monitoring. Accurate identification of mature fruit (Stage G) and semi-mature stages (Stage F) allows growers or autonomous systems to time harvest operations precisely, maximizing fruit quality and minimizing losses due to over-ripeness [83]. These models can be embedded into robotic harvesting platforms (as has been demonstrated for other crops; [84–87]), that operate continuously and non-destructively in production systems. Moreover, by combining the Raspberry PhenoSet-trained YOLOv8 model with the Neural Radiance Fields (NeRFs) [88], it is possible to construct 3D maps of plant canopies that encode the spatial location of buds, flowers, and fruits across the farm. Such maps can support path optimization, selective harvesting, and crop load balancing in densely planted environments like vertical farms or polytunnels.

Third, Raspberry PhenoSet could be extended for use in plant health monitoring. Because the dataset was collected in a pesticide-free vertical farming facility, it represents a clean baseline for phenological detection. By augmenting the dataset with additional images or annotations of disease symptoms, nutrient imbalances, or pollination deficits, it may be adapted to detect early signs of plant stress. This would enable low-cost surveillance of physiological health using the same infrastructure as phenology monitoring, reducing the need for separate sensing systems [89–91].

Finally, the dataset and benchmarked models provide a foundation for future research in multimodal sensing. The challenges observed in distinguishing certain developmental stages, especially in visually cluttered environments, could be mitigated through the integration of spectral or thermal features, as shown in prior work on flower detection using near-infrared and multispectral imaging [14, 92, 93]. Further model development using expanded training data from diverse lighting conditions, cultivars, and cropping systems would increase generalizability and support broader commercialization of phenology-based automation tools. Moreover, the established dataset can be used for deep learning model development.

# 4 Conclusion

This study addressed the critical challenge of precise harvest timing for highly perishable crops such as raspberries by developing an integrated phenology-based computer vision framework. To this end, a phenology study of three raspberry cultivars was carried out, and seven phenology stages were identified based on the BBCH scale. Then, a phenology-based dataset of raspberries was curated and made publicly available for the first time. This resource enabled the development and comprehensive benchmarking of deep learning models, from which a tailored YOLOv8-x model emerged as optimal, achieving a 92.2% detection accuracy in field validation and real-time operation. The dataset and models were also validated under realistic deployment conditions, with the model operating at frame rates compatible with real-time applications.

The practical significance of this work lies in its ability to provide growers with a reliable tool for advancing yield forecasting by 28–33 days, a capability crucial for optimizing labour resources, fulfilling contractual obligations, and reducing postharvest waste. By demonstrating robust performance on consumer-grade hardware in an operational vertical farm environment, this research confirms the feasibility of deploying such deep learning computer vision solutions at scale. Given the global market value of raspberries and the increasing need for automation in horticulture, phenology-based detection frameworks such as those demonstrated here represent a timely and valuable contribution to the agri-food technology sector.

## Appendix I: Assessment of Preprocessing Techniques on Performance

To further investigate preprocessing and architectural modifications, additional experiments were performed. These additional experiments focus on (1) background contrast enhancement and (2) shallow feature fusion for improved small-target detection.

### Contrast enhancement via CLAHE

To assess the impact of image preprocessing on detection performance, Contrast Limited Adaptive Histogram Equalization (CLAHE) was applied to the luminance channel of all images in the dataset. The enhanced dataset was used to retrain the detector under identical training settings (same architecture, data splits, batch size, and training schedule) to ensure a fair comparison.

The performance of the baseline model was comparable with the performance of the model with CLAHE contrast enhancement on images as shown in Table 7. This can be attributed to the fact that the images were captured under controlled indoor conditions with sufficient illumination and contrast. As a result, additional contrast enhancement does not provide new information.

Table 7 - Performance of YOLOv8-x on Contrast-Enhanced Raspberry PhenoSet

| Class | Precision | | Recall | | AP50 | | F1-Score | |
|---|---|---|---|---|---|---|---|---|
| | Baseline | CLAHE | Baseline | CLAHE | Baseline | CLAHE | Baseline | CLAHE |
| A | 0.802 | 0.782 | 0.819 | 0.798 | 0.855 | 0.837 | 0.810 | 0.790 |
| B | 0.773 | 0.668 | 0.635 | 0.687 | 0.729 | 0.708 | 0.697 | 0.677 |
| C | 0.621 | 0.570 | 0.534 | 0.579 | 0.592 | 0.583 | 0.574 | 0.574 |
| D | 0.641 | 0.595 | 0.652 | 0.696 | 0.677 | 0.647 | 0.646 | 0.642 |
| E | 0.731 | 0.642 | 0.569 | 0.582 | 0.640 | 0.654 | 0.640 | 0.611 |
| F | 0.849 | 0.787 | 0.577 | 0.586 | 0.678 | 0.655 | 0.687 | 0.672 |
| G | 0.632 | 0.719 | 0.891 | 0.854 | 0.850 | 0.851 | 0.739 | 0.781 |
| **Mean** | 0.721 | 0.681 | 0.668 | 0.683 | 0.717 | 0.705 | 0.693 | 0.682 |

**Shallow Feature Fusion (SFF)**

To evaluate whether enhancing high-resolution feature propagation improves the detection of small or early-stage targets, a modified detection architecture was implemented by introducing an additional shallow feature fusion branch. To this end, a higher-resolution feature map from an earlier backbone stage (P2 level) was fused into the detection head using up-sampling and concatenation operations, extending the standard multi-scale detection structure.

The modified model was trained under the same conditions as the baseline to ensure a controlled comparison. The shallow feature fusion did not result in additional improvement in the performance of the baseline model (Table 8). This is likely because the baseline detector already incorporates effective multi-scale feature aggregation, and the target objects are sufficiently represented at existing feature resolutions.

Table 8 - Performance of Modified YOLOv8-x for Shallow Feature Fusion on Raspberry PhenoSet

| Class | Precision | | Recall | | AP50 | | F1-Score | |
|---|---|---|---|---|---|---|---|---|
| | Baseline | SFF | Baseline | SFF | Baseline | SFF | Baseline | SFF |
| A | 0.802 | 0.852 | 0.819 | 0.766 | 0.855 | 0.850 | 0.810 | 0.807 |
| B | 0.773 | 0.811 | 0.635 | 0.626 | 0.729 | 0.741 | 0.697 | 0.707 |
| C | 0.621 | 0.686 | 0.534 | 0.555 | 0.592 | 0.603 | 0.574 | 0.614 |
| D | 0.641 | 0.652 | 0.652 | 0.677 | 0.677 | 0.639 | 0.646 | 0.664 |
| E | 0.731 | 0.684 | 0.569 | 0.580 | 0.640 | 0.598 | 0.640 | 0.628 |
| F | 0.849 | 0.853 | 0.577 | 0.529 | 0.678 | 0.657 | 0.687 | 0.653 |
| G | 0.632 | 0.786 | 0.891 | 0.907 | 0.850 | 0.886 | 0.739 | 0.842 |
| **Mean** | 0.721 | 0.761 | 0.668 | 0.663 | 0.717 | 0.710 | 0.693 | 0.709 |

## CRediT authorship contribution statement

**Parham Jafary:** Conceptualization, Data curation, Formal analysis, Investigation, Methodology, Software, Validation, Visualization, Writing – original draft, Writing – review & editing. **Lesley G. Campbell:** Conceptualization, Formal analysis, Funding acquisition, Methodology, Project administration, Resources, Supervision, Writing – original draft, Writing – review & editing. **Anna Bazangeya:** Conceptualization, Data curation, Formal analysis, Investigation, Visualization, Writing – review & editing. **Michelle Pham:** Conceptualization, Data curation, Formal analysis, Investigation, Visualization, Writing – review & editing**. Sajad Saeedi:** Supervision, Writing – review & editing. **Kourosh Zareinia:** Supervision, Funding acquisition, Writing – review & editing**. Habiba Bougherara:** Conceptualization, Funding acquisition, Resources, Supervision, Writing – review & editing.

## Data availability statement

Data is publicly available at https://sites.google.com/view/raspberry-phenoset.

## Funding statement

This work was supported by Weston Family Foundation via Homegrown Innovation Challenge [grant number: SA-10691 to LGC and HB], and NSERC Grants [RGPIN-2019-05780 to LGC; USRA to AB, RGPIN-2019-05615 to HB, RGPIN-2019-05562 to KZ]

## Statement of competing interests

Authors declare that they have no known competing interests.